\pdfoutput=1

\documentclass[11pt,a4paper]{article}
\usepackage[hyperref]{swisstext2021}
\usepackage{times}
\usepackage{latexsym}
\usepackage{todonotes}
\usepackage{numprint}
\usepackage{tabularx}
\usepackage{booktabs}

\usepackage{microtype}

\usepackage{url}

\aclfinalcopy 

\title{Swiss Parliaments Corpus, an Automatically Aligned Swiss German Speech to Standard German Text Corpus}

\author{Michel Pl{\"u}ss\qquad Lukas Neukom\qquad Christian Scheller\qquad Manfred Vogel \\
Institute for Data Science \\
University of Applied Sciences and Arts Northwestern Switzerland \\
Windisch, Switzerland \\
\texttt{michel.pluess@fhnw.ch} \\}

\begin{document}
\maketitle
\begin{abstract}

We present the Swiss Parliaments Corpus (SPC), an automatically aligned Swiss German speech to Standard German text corpus. This first version of the corpus is based on publicly available data of the Bernese cantonal parliament and consists of 293 hours of data. It was created using a novel forced sentence alignment procedure and an alignment quality estimator, which can be used to trade off corpus size and quality. We trained Automatic Speech Recognition (ASR) models as baselines on different subsets of the data and achieved a Word Error Rate (WER) of 0.278 and a BLEU score of 0.586 on the SPC test set. 
The corpus is freely available for download\footnote{\url{https://www.cs.technik.fhnw.ch/i4ds-datasets}}.

\end{abstract}

\section{Introduction}

Swiss German is a family of dialects spoken by around five million people in Switzerland. It is different from Standard German regarding phonetics, vocabulary, morphology, and syntax. Swiss German is mostly a spoken language. While it is also used in writing, particularly in informal text messages, it lacks a standardized writing system. This leads to difficulties for automated text processing such as spelling ambiguities and a huge vocabulary size. For Swiss German ASR, we therefore focus on end-to-end approaches from Swiss German speech to Standard German text. This can be viewed as a Speech Translation problem with similar source and target languages. For example, the Swiss German sentence "Ide Abfahrt hetter de sächsti Platz beleit" can be translated to the
Standard German sentence "In der Abfahrt belegte er den sechsten Platz". Here, the past tense changes.

Currently, training an ASR model for Swiss German is challenging due to the lack of public training data. Only a few hours of Swiss German speech with Standard German text are available. To reach high-quality ASR results, a corpus with thousands of hours of transcribed speech is required. For example, \citet{librispeechsoa} set the current state-of-the-art on the English LibriSpeech~\cite{librispeech} test-other benchmark with a WER of 0.034 using 960 hours of labeled training data and another 57700 hours of unlabeled data.

While there is no ready-to-use training data, many Swiss parliaments record their debates. Most communal and some cantonal parliaments hold their meetings in Swiss German. Some of them do a full transcript of the recordings in Standard German resulting in more than 4000 hours of raw data.

To transform the raw data into training data, we developed a novel forced sentence alignment algorithm which handles the problems created by the language mismatch between audio and text such as changes to the word order within a sentence (sentence reordering). It is based on a German ASR model and global alignment and includes a learned filter component specifically tuned for the Swiss German speech to Standard German text use case\footnote{The code is available in our GitHub repository:  \url{https://github.com/festivalhopper/swiss-parliaments-corpus-paper}}. Using the developed alignment algorithm, we created and published a first corpus called the Swiss Parliaments Corpus consisting of data from the parliament Grosser Rat Kanton Bern.

The remainder of this paper is structured as follows: Related work is discussed in section 2. The forced sentence alignment procedure is described in section 3. Details about our corpus can be found in section 4. Section 5 contains baseline models and experiments. Section 6 wraps up the paper and gives directions for future work.

\section{Related Work}

An earlier version of this corpus was previously published as part of the Low-Resource Speech-to-Text shared task at GermEval 2020 (GermEval Task 4)~\cite{germeval2020task4}. To our knowledge, there are only two other publicly available corpora for Swiss German ASR. ArchiMob~\cite{archimob} includes 69 hours of Swiss German speech and corresponding Swiss German transcripts. Unfortunately, Standard German transcripts are not available. The Radio Rottu Oberwallis dataset~\cite{radio-rottu} includes 8 hours of speech, of which only 2 hours have Standard German transcripts in addition to Swiss German transcripts.
Furthermore, the Standard German dataset of the Common Voice~\cite{ardila2019common} ASR corpus has $1$ \% of its utterances spoken in a Swiss German accent, which however strongly differs from actual Swiss German speech.

There are different approaches to forced alignment of long speech recordings in the context of creating an ASR corpus. 
Our procedure is similar to that described in \citet{Hazen2006AutomaticAA,librispeech,pratap2020mls}.
Like our approach, these methods initially transcribe audios using an ASR system, followed by an alignment stage and a final refinement stage.
The main differences are that these approaches do not yield strictly sentence-level alignments, which are a requirement for our work due to the possibility of sentence reorderings between Standard German and Swiss German, and our novel approach to filter the corpus and improve the quality.

\section{Forced Sentence Alignment Procedure}

Our forced sentence alignment procedure takes a Swiss German recording of arbitrary length and the corresponding manual Standard German transcript as inputs. The audio file is transcribed with an ASR model. An important requirement for this model is the ability to annotate accurate start and end times of each word in the output. Since no publicly available Swiss German ASR model with this feature exists, we resort to a Standard German model.
The ASR transcript is then globally aligned to the manual transcript using the Biopython~\cite{biopython} implementation of the Needleman-Wunsch algorithm~\cite{needleman-wunsch}. The manual transcript is split into sentences using spaCy~\cite{spacy2}. Each of these sentences is mapped to a start and end time in the recording via the global alignment and the per-word start and end times provided by the ASR model.

\subsection{Alignment Corpus and Metrics}

We created a separate internal alignment corpus to be able to measure the quality of our sentence alignment. It consists of almost 6 hours of transcribed recordings from four different parliaments and one other data source. We split the corpus into a training and a test set (60-40 split). Recordings and transcripts were manually sentence-aligned.

We define an aligned sentence as a three-tuple of the sentence, its start and end time. We call an aligned sentence empty if the start and end times are not set, which means the sentence is not spoken in the recording. This happens because transcripts sometimes have errors such as missing or additional sentences.

Our goal is to maximize the following metrics during the creation of the corpus:

\begin{itemize}
\item \textbf{The Intersection over Union (IoU)} reflects the alignment quality. We report the mean IoU over all predicted aligned sentences for which the manual as well as the predicted aligned sentence are not empty. 
\item \textbf{The sentence precision and recall} reflect the corpus quality and size. A predicted aligned sentence counts as true positive (TP) if the manual as well as the predicted aligned sentence are not empty. True negative (TN) means the manual as well as the predicted aligned sentence are empty. False positive (FP) means the manual aligned sentence is empty, but the predicted aligned sentence is not empty. False negative (FN) means the manual aligned sentence is not empty, but the predicted aligned sentence is empty. The sentence precision is equal to TP / (TP + FP). The sentence recall is equal to TP / (TP + FN).
\end{itemize}
\subsection{IoU Estimate Filter and Further Refinements}

We filter out sentences with a bad alignment quality based on an estimate of their IoU. We fit a Gradient Boosting regressor to estimate a sentence's IoU using the following features:

\begin{itemize}
\item \textbf{Length ratio} of the manual transcript sentence to the part of the ASR transcript it was aligned to
\item \textbf{Alignment score} of the manual transcript sentence, normalized by its length
\item \textbf{Mean speech recognition confidence} as reported by the ASR system over the words the manual transcript sentence was aligned to
\item \textbf{Chars per second}, i.e. ratio of the manual transcript sentence length to the audio length (predicted aligned sentence end time minus start time)
\end{itemize}

We use the LightGBM implementation by Ke et al.~\shortcite{lightgbm}. Table~\ref{lightgbm-hyperparams} shows our hyperparameters. These were found using Bayesian optimization.

\begin{table}
\begin{center}
\npdecimalsign{.}
\nprounddigits{3}
\begin{tabular}{lc}
\toprule \textbf{Hyperparameter} & \textbf{Value} \\ \midrule
num\_leaves & 3 \\
min\_child\_samples & 7 \\
max\_bin & 7597 \\
\bottomrule
\end{tabular}
\end{center}
\caption{\label{lightgbm-hyperparams} LightGBM hyperparameters for the IoU regressor. For all other parameters we use the default values as defined by the LightGBM authors.}
\end{table}

The regressor estimates the IoU in a 3-fold cross validation experiment on the training set of our alignment corpus with a mean absolute error of 0.108 (IoU values are in the interval [0, 1]). We propose two different IoU estimate thresholds. A threshold of 0.7 is supposed to keep as many sentences as possible and only discard sentences with a bad alignment quality, e.g. for a training set. A threshold of 0.9 is supposed to keep only sentences with a very good alignment quality, e.g. for a test set. Thresholds were found using a parameter sweep on the test set of the alignment corpus.

Two more refinements were implemented: to filter out manual transcripts that are clearly mismatched or incomplete, no alignment is created if the length ratio of the longer transcript to the shorter transcript is greater than six. The optimal ratio was found using a parameter sweep on the alignment corpus test set.
Finally, we fit a start and and end time correction offset on the training set of our alignment corpus and calibrate the start and end times of each sentence by adding the correction offset. This leads to a minor IoU improvement because the times reported by the ASR model can be slightly off.

\subsection{Experiments and Results}

\begin{table}
\begin{center}
\npdecimalsign{.}
\nprounddigits{3}
\begin{tabular}{ln{1}{3}}
\toprule \textbf{ASR Model} & \textbf{WER} \\ \midrule
Amazon Transcribe & 0.6258 \\
Google Speech-to-Text & 0.7250 \\
\bottomrule
\end{tabular}
\end{center}
\caption{\label{stt-comparison} Comparison of the performance of different ASR models on the GermEval 2020 Task 4 public test set }
\end{table}

\begin{table*}[t!]
\begin{center}
\npdecimalsign{.}
\nprounddigits{3}
\begin{tabular}{lp{0.315\linewidth}n{1}{3}n{1}{3}n{1}{3}n{1}{3}n{1}{3}n{1}{3}}
\toprule \textbf{ASR Model} & \textbf{Settings} & \begin{tabular}{@{}l@{}}\textbf{Mean} \\ \textbf{IoU}\end{tabular} & \begin{tabular}{@{}l@{}}\textbf{Sentence} \\ \textbf{Precision}\end{tabular}\textbf{ } & \begin{tabular}{@{}l@{}} \textbf{Sentence} \\ \textbf{Recall}\end{tabular} \\ \midrule
Amazon Transcribe & No Refinements & 0.8318 & 1.0000 & 0.9555 \\
Amazon Transcribe & Length Ratio & 0.8360 & 1.0000 & 0.9491 \\
Amazon Transcribe & Time Calibration & 0.8360 & 1.0000 & 0.9555 \\
Amazon Transcribe & Length Ratio + Time Calibration & 0.8401 & 1.0000 & 0.9491 \\
Google Speech-to-Text & Length Ratio + Time Calibration & 0.6889 & 1.0000 & 0.9936 \\
Amazon Transcribe & Length Ratio + Time Calibration + IoU Estimate Filter 0.7 & 0.8883 & 1.0000 & 0.8219 \\
Amazon Transcribe & Length Ratio + Time Calibration + IoU Estimate Filter 0.9 & 0.9271 & 1.0000 & 0.4881 \\ 
\bottomrule
\end{tabular}
\end{center}
\caption{\label{alignment-results} Sentence alignment metrics on the test set of our alignment corpus for two ASR models with various settings }
\end{table*}

We evaluated two ASR models, Amazon Transcribe\footnote{\url{https://aws.amazon.com/transcribe}}\footnote{We used the "Swiss German" model. Based on the information we found, we believe this is a model for Standard German, specialized on Swiss accents, not for actual Swiss German.} and Google Speech-to-Text\footnote{\url{https://cloud.google.com/speech-to-text}}\footnote{We used the "German (Germany)" model, "German (Switzerland)" was not yet available at the time of the experiment.}, on the public test set of GermEval 2020 Task 4~\cite{germeval2020task4}. Table~\ref{stt-comparison} shows the results of this comparison. Amazon is ahead of Google by 0.1 WER. This suggests that Amazon succeeded in improving the performance for Swiss German with its specialized model, but still has a long way to go to achieve a general-purpose ASR model with a WER comparable to English or Standard German models. In comparison to the performance of the winning contribution to GermEval 2020 Task 4 by Büchi et al.~\shortcite{zhaw2020} with a WER of 0.403, Amazon Transcribe is more than 0.2 WER behind.

We conducted experiments using Amazon Transcribe and Google Speech-to-Text as the ASR engines with different combinations of refinements\footnote{We could not use the Büchi et al. model because it does not provide word start and end times}. We determined the parameters for the global alignment algorithm using Bayesian optimization with 3-fold cross validation on the training set of our alignment corpus. They are listed in appendix~\ref{alignment-parameters-optimized}. Table~\ref{alignment-results} shows the results. The lead in WER by 0.1 (see table~\ref{stt-comparison}) for Amazon translates to a mean IoU that is 0.151 higher than Google's result with the same settings. The 0.045 advantage of the latter in sentence recall does not make up for this, even less so because we prefer quality over quantity. For Amazon Transcribe, enabling length ratio filtering as well as time calibration appears to be the best option, resulting in a mean IoU of 0.840 and a sentence recall of 0.949. The alignment quality can be further improved using the IoU estimate filter. A threshold of 0.7 leads to an increase of 0.048 in mean IoU and a decrease of 0.127 in sentence recall, whereas a threshold of 0.9 leads to an increase of 0.087 in mean IoU and a decrease of 0.461 in sentence recall. The sentence precision is perfect in all experiments.

\section{Swiss Parliaments Corpus}

Using our forced sentence alignment procedure, we created and published\footnote{\url{https://www.cs.technik.fhnw.ch/i4ds-datasets}} a corpus called the Swiss Parliaments Corpus. It is based on recordings and transcripts from the parliament Grosser Rat Kanton Bern\footnote{\url{https://www.gr.be.ch/gr/de/index/sessionen/sessionen.html}}. As expected, given the location of the parliament, most speakers have a Bernese dialect. The recordings are MP4 videos, one video per parliament meeting, with a length spanning from 28 minutes to 4 hours and 2 minutes. The transcripts are in PDF format, with one PDF containing a whole session with usually around 10 to 15 meetings.

\subsection{Corpus Parts}

\begin{table}
\begin{center}
\npdecimalsign{.}
\nprounddigits{3}
\begin{tabular}{l l l}
\toprule \textbf{Corpus Part} & \begin{tabular}{@{}l@{}}\textbf{Audio Length} \\ \textbf{in Hours}\end{tabular} & \begin{tabular}{@{}l@{}}\textbf{Number of} \\ \textbf{Speakers}\end{tabular}\\ \midrule
Raw data & 460 & - \\
train\_all & 293 & 198 \\
train\_0.7 & 256 & 195 \\
train\_0.9 & 176 & 194 \\
test & 6 & 26 \\
\bottomrule
\end{tabular}
\end{center}
\caption{\label{corpus-part-sizes} Overview of the different subsets of the corpus, their sizes and the number of unique speakers }
\end{table}

Table~\ref{corpus-part-sizes} gives an overview of the different corpus parts and their sizes. We created an unfiltered training set called train\_all with 293 hours of data. We then used IoU estimate filtering to create two training subsets, train\_0.7 with a threshold of 0.7 and 256 hours of data as well as train\_0.9 with a threshold of 0.9 and 176 hours of data. The unfiltered training set contains an IoU estimate column to create a training set with a custom threshold. The test set was created with a threshold of 0.9 and contains 6 hours of data. We could therefore transform 65 \% of the raw data to training or test data.

\subsection{Settings, Filters, Split}

We used semi-global alignment parameters (no gap penalties on the start and end of both sequences, see appendix~\ref{alignment-parameters-semi-global}) to deal with incomplete recordings and additional irrelevant recorded audio. Length ratio filtering was disabled while time calibration was enabled. We applied the following additional filters:
\begin{itemize}
\item Chars per second must be between 6 and 23. The average for chars per second in this corpus is 15. Aligned sentences outside of this range probably either contain a lot of idle time in the recording or additional text that is not recorded.
\item We detect the language of each sentence using langdetect\footnote{\url{https://pypi.org/project/langdetect}} and only keep German sentences.
\item (Test set only) Audio length must be at least 1 second.
\item (Test set only) Audio length must be less than 15 seconds.
\item (Test set only) Sentences must be unique across the whole dataset.
\end{itemize}
Speakers are automatically deduplicated. The train-test split guarantees that the utterances of a speaker are only contained in either the training set or the test set, never in both. To ensure that the speakers in the test set are diverse enough, a speaker can only be part of the test set if her or his utterances make up less than 10 \% of the whole test set.

\section{ASR Baselines}

All the baseline models are implemented using the ESPnet framework \cite{watanabe2018espnet}. We trained a Transformer model \cite{vaswani2017attention} as well as a Conformer model \cite{gulati2020conformer}. The network architectures closely follow the Common Voice example of ESPnet\footnote{\url{https://github.com/espnet/espnet/tree/master/egs2/commonvoice/asr1}} using a hybrid CTC/attention encoder-decoder framework \cite{watanabe2017hybrid}. 

Inputs are first down-sampled to 1/4 length by two strided 2D convolution layers and ReLU activations. 
The Transformer encoder consists of 12 self-attention blocks with 2048 units. 
Similarly, the Conformer encoder uses 12 Conformer-layers with 2048 units. 
Both Transformer and Conformer models use a Transformer decoder with six self-attention blocks with 2048 units.

For the training of both models we use the Adam optimizer~\citep{kingma2015adam} and a warmup learning rate schedule similar to the one proposed in~\citet{vaswani2017attention}, but with a fixed warmup period of 25000 steps and a maximum learning rate of 0.002. 
As input we use 80-channel log-mel filterbanks that are shifted to have a mean of zero. 
Speed perturbation~\citep{ko2015audio} with random factors between 0.9 and 1.1 and SpecAugment \cite{park2019specaugment} are used for data augmentation.
Both models are trained with a combined Connectionist Temporal Classification (CTC)~\citep{graves2006connectionist} and cross entropy loss with weights 0.3 and 0.7, respectively.

During decoding we use a 16-layer Transformer language model, trained on the SPC texts as well as the German texts of the EuroParl Corpus~\cite{koehn2005europarl}, using beam search with a beam size of 50.

We trained both models on all subsets of the SPC until convergence for 200 epochs. The results of all models are shown in Table \ref{experiment-results}. We report WER, commonly used to evaluate ASR systems, as well as the BLEU~\citep{papineni2002bleu} score, commonly used to evaluate Machine Translation and Speech Translation systems. For BLEU, we use the implementation provided by NLTK~\citep{bird2009natural} with default parameters. In our experiments, WER and BLEU show a negative correlation as expected (WER: lower is better, BLEU: higher is better), indicating that both are similarly useful metrics.

The Conformer model performs better than the Transformer model in all experiments. This is in line with the findings of \citet{gulati2020conformer} on the LibriSpeech benchmark. For both models, all dataset splits resulted in similar WER and BLEU scores, even though their sizes differ significantly (see Table~\ref{corpus-part-sizes}). This can be explained by the higher quality when filtering based on IoU scores. As a result, we can train models on 60 \% of the data, resulting in faster training, without losing performance.

\begin{table}
\begin{center}
\npdecimalsign{.}
\nprounddigits{3}
\begin{tabular}{lln{1}{3}n{1}{3}}
\toprule \textbf{Model} & \textbf{Dataset}& \textbf{WER} & \textbf{BLEU} \\ \midrule
Transformer & train\_all & 0.2968676658463045 & 0.5478920956667053 \\
 & train\_0.7 & 0.2928272325879641 & 0.5525188279890423 \\
 & train\_0.9 & 0.3029852102807204 & 0.5374058427024371 \\
\midrule
Conformer & train\_all & 0.289300474826851 & 0.57712026864062 \\
 & train\_0.7 & 0.278428583367282 & 0.58563332211857 \\
 & train\_0.9 & 0.286931601153651 & 0.576848096660676 \\
\bottomrule
\end{tabular}
\end{center}
\caption{\label{experiment-results} Test WER and BLEU scores of Transformer and Conformer models on all subsets of the SPC.}
\end{table}

\section{Conclusion}

In this work, we introduced the Swiss Parliament Corpus, an automatically aligned Swiss German speech to Standard German text corpus.
We proposed a multi-stage forced sentence alignment procedure that leverages existing Standard German ASR systems and uses a novel IoU estimator for refinement.
We also provided Transformer and Conformer ASR baseline models that showcase the benefits of the IoU estimates. The best model achieves a WER of 0.278 and a BLEU score of 0.586 on the SPC test set.

We believe that our forced sentence alignment procedure is a step towards making large-vocabulary speech recognition for all Swiss German dialects possible. The SPC with its 293 hours of training data supports this thesis. It is freely available for download\footnote{\url{https://www.cs.technik.fhnw.ch/i4ds-datasets}}.

In future work, we plan to increase the corpus size and the dialect diversity by aligning recordings and transcripts of additional parliaments. We also plan to collect data for a test set representing all Swiss German dialects since the SPC is domain-specific and includes mostly Bernese speakers. This would facilitate a fair comparison for Swiss German ASR systems. In the future, we will also use the trained models to improve the forced sentence alignment algorithm results in a similar way as \citet{sennrich-volk-2011-iterative}.

Furthermore, the effects of the IoU filter on the quality of the dataset as well as the impact on the model need further investigation.
Finally, we want to investigate the correlation of WER and BLEU with human evaluation to understand which metric is most appropriate for the problem. In this context, it would also be interesting to further investigate the differences between a literal transcription in Swiss German and a Standard German translation.

\section*{Acknowledgments}

First and foremost, we would like to thank the parliamentary services of the canton of Bern for their work on the transcription of the debates and for publishing recordings and transcripts on their website. Without them, the SPC would not exist.

Furthermore, we thank Pascal Thormeier who created the first version of our alignment corpus during his bachelor's thesis.

We also thank the participants of GermEval 2020 Task 4 for the fruitful discussions in the aftermath of the task, which lead to several improvements of the SPC.

\bibliography{swiss_parliaments_corpus}
\bibliographystyle{acl_natbib}

\appendix
\section{Alignment Parameters Optimized on Alignment Corpus}
\label{alignment-parameters-optimized}

\begin{table}[h]
\begin{center}
\npdecimalsign{.}
\nprounddigits{3}
\begin{tabular}{ln{1}{3}}
\toprule \textbf{Alignment Parameter} & \textbf{Value} \\ \midrule
match\_score & 0.03875752471676385 \\
mismatch\_score & -1.0 \\
truth\_left\_open\_gap\_score & -0.5038367052042227 \\
truth\_internal\_open\_gap\_score & -1.0 \\
truth\_right\_open\_gap\_score & -0.43980186690399603 \\
truth\_left\_extend\_gap\_score & -0.2440180768676541 \\
truth\_internal\_extend\_gap\_score & -0.4817146150129493 \\
truth\_right\_extend\_gap\_score & -0.2594102766979399 \\
stt\_left\_open\_gap\_score & -1.0 \\
stt\_internal\_open\_gap\_score & -0.7698209478188247 \\
stt\_right\_open\_gap\_score & -0.9815365376036425 \\
stt\_left\_extend\_gap\_score & -0.25266456311369834 \\
stt\_internal\_extend\_gap\_score & -0.7698209478188247 \\
stt\_right\_extend\_gap\_score & -0.5619337177636895 \\
\bottomrule
\end{tabular}
\end{center}
\caption{\label{alignment-parameters-optimized-table} Alignment parameters found using Bayesian optimization with 3-fold cross validation on the training set of our alignment corpus}
\end{table}

\pagebreak

\section{Alignment Parameters for SPC}
\label{alignment-parameters-semi-global}

\begin{table}[h]
\begin{center}
\begin{tabular}{lc}
\toprule \textbf{Alignment Parameter} & \textbf{Value} \\ \midrule
match\_score & 1.0 \\
mismatch\_score & -1.0 \\
truth\_left\_open\_gap\_score & 0.0 \\
truth\_internal\_open\_gap\_score & -1.0 \\
truth\_right\_open\_gap\_score & 0.0 \\
truth\_left\_extend\_gap\_score & 0.0 \\
truth\_internal\_extend\_gap\_score & -1.0 \\
truth\_right\_extend\_gap\_score & 0.0 \\
stt\_left\_open\_gap\_score & 0.0 \\
stt\_internal\_open\_gap\_score & -1.0 \\
stt\_right\_open\_gap\_score & 0.0 \\
stt\_left\_extend\_gap\_score & 0.0 \\
stt\_internal\_extend\_gap\_score & -1.0 \\
stt\_right\_extend\_gap\_score & 0.0 \\
\bottomrule
\end{tabular}
\end{center}
\caption{\label{alignment-parameters-semi-global-table} Alignment parameters used to create the SPC}
\end{table}

\end{document}